\documentclass[11pt,a4paper]{article}
\pdfoutput=1
\usepackage{times}
\usepackage{latexsym}
\usepackage{url}
\usepackage[T1]{fontenc}

\usepackage{amsmath}
\usepackage{booktabs}
\usepackage{multirow}
\usepackage{graphicx}

\usepackage[acceptedWithA]{tacl2018v2}

\usepackage{xspace,mfirstuc,tabulary}

\newif\iftaclinstructions
\taclinstructionsfalse 
\iftaclinstructions
\renewcommand{\confidential}{}
\renewcommand{\anonsubtext}{(No author info supplied here, for consistency with
TACL-submission anonymization requirements)}
\newcommand{\instr}
\fi

\iftaclpubformat 

\else

\fi

\title{A Knowledge-Enhanced Pretraining Model for Commonsense Story Generation}

\author{
 Jian Guan$^1$, Fei Huang$^1$, Zhihao Zhao$^2$, Xiaoyan Zhu$^1$, Minlie Huang$^1$\Thanks{Corresponding author: Minlie Huang.}\\
 $^1$Institute for Artificial Intelligence, State Key Lab of Intelligent Technology and Systems \\
 $^1$Beijing National Research Center for Information Science and Technology\\
 $^1$Department of Computer Science and Technology, Tsinghua University, Beijing 100084, China\\
 $^2$School of Software, Beihang University, Beijing, China\\
  {\sf j-guan19@mails.tsinghua.edu.cn},{\sf f-huang18@mails.tsinghua.edu.cn}, \\{\sf extsuioku@gmail.com}, {\sf zxy-dcs@tsinghua.edu.cn}, {\sf aihuang@tsinghua.edu.cn} \\
}

\date{}

\begin{document}
\maketitle
\begin{abstract}
Story generation, namely generating a reasonable story from a leading context, is an important but challenging task. In spite of the success in modeling fluency and local coherence, existing neural language generation models (e.g., GPT-2) still suffer from repetition, logic conflicts, and lack of long-range coherence in generated stories. We conjecture that this is because of the difficulty of associating relevant commonsense knowledge, understanding the causal relationships, and planning entities and events with proper temporal order. 
In this paper, we devise a knowledge-enhanced pretraining model for commonsense story generation. We propose to utilize commonsense knowledge from external knowledge bases to generate reasonable stories. To further capture the causal and temporal dependencies between the sentences in a reasonable story, we employ multi-task learning which combines a discriminative objective to distinguish true and fake stories during fine-tuning. Automatic and manual evaluation shows that our model can generate more reasonable stories than state-of-the-art baselines, particularly in terms of logic and global coherence.


\end{abstract}

\section{Introduction}

\begin{table}[!ht]
\scriptsize
\centering
\begin{tabular}{p{7.2cm}}
\toprule
\textsf{\textbf{Context:}} \\ 
\textsf{The first time I saw the results of an \textbf{accident}, I was in a {bus}.}\\
\midrule
\textsf{\textbf{Golden Continuation by Human:}}\\ 
\textsf{We were {stopped} and a friend pointed out the accident on the road. A man was under a car after having {been run over}. People tried to {move the car} off him but only made it worse. After seeing that, I {am much more careful} when {crossing the street}.} \\\\

\textsf{\textbf{pretrained GPT-2:}} \\
\textsf{\textit{She was trying to get home but the other passengers would not let her go}. He thought she \underline{was going to die because of her weight}. \textit{She was trying to get home but the other passengers would not let her go}. The bus driver \textit{{would not let her go}}.}\\\\
\textsf{\textbf{Fine-tuned GPT-2:}} \\
\textsf{\textit{I was on my way to a party}. \textit{I was on my way to a party} and I 'd gotten out of my seat, and started driving. I got a flat tire, so I \underline{stopped driving}. I \underline{drove to the party} and had a great time.}\\

\bottomrule
\end{tabular}
\caption{Story examples generated by human and GPT-2 models. The stories written by the pretrained GPT-2 and fine-tuned GPT-2 (post-trained on ROCStories~\cite{Mostafazadeh2016Story}) suffer from repetition~(in \textit{italic}), bad inter-sentence coherence to the context~(e.g., ignoring key entities such as \texttt{accident} in \textbf{bold}), as well as conflicting logic~(\underline{underlined}, e.g., first \texttt{stopped driving} and then \texttt{drove to the party}), in spite of their good fluency and intra-sentence coherence.
}
\label{intro}
\end{table}

Story generation is a strong indicator of machine understanding of natural language. It is often approached as selecting a sequence of events to form a story with a reasonable logic or plot. While existing generative models~\cite{roemmele2016seq2seq,fan2018hierarchical,fan-etal-2019-strategies} can generate stories with good local coherence, they are still struggling to plan a coherent plot and maintain a reasonable event sequence throughout the story, or they are often biased towards generating a limited set of stories with generic plots~\cite{see2019massively} (e.g., \texttt{I have a great time}), even when using the powerful generative model OpenAI's GPT-2~\cite{radford2019language}, as shown in Table~\ref{intro}.  


{Pretrained GPT-2 has been shown to capture useful semantic and syntactic features~\cite{alt-etal-2019-fine}, 
as demonstrated by state-of-the-art performance on some generation tasks such as machine translation 
and text summarization~\cite{radford2019language}. However, compared with such tasks whose source inputs have contained sufficient information to generate desired target texts, 
story generation is a typical \emph{open-ended generation} task, where only very limited information is given in the input. 
As shown in this paper, we observe some severe issues when applying GPT-2 to generate reasonable stories, particularly commonsense stories from a limited beginning. 
These issues include \emph{repetition, logic conflicts, and lack of long-range coherence}~\cite{see2019massively,holtzman2019curious}, as exemplified in Table~\ref{intro}. Specifically, although GPT-2 performs reasonably well at generating some related concepts to \texttt{bus}~(e.g., \texttt{driver}, and the probable destinations \texttt{home} or \texttt{party}), it completely ignores the other key entity \texttt{accident} in the leading context, which could be caused by its lower frequency in GPT-2's initial training corpus~(less than 7\% of \texttt{bus}). Besides, even though the concepts are relevant, they are usually generic, and used repeatedly and illogically in the generated stories. Therefore, 
given limited information as input, it is extremely challenging for the subsequent generation without any external guidance, for instance, commonsense knowledge. And the difficulties lie in associating inter-dependent commonsense knowledge for expanding a reasonable story, handling the causal relationships, as well as deciding the temporal 
orders between entities and events in context.} 



{Explicitly introducing external commonsense knowledge
has been shown helpful 
to improve language understanding and  long-range coherence of generated texts~\cite{Zhou2018Commonsense,guan2019story,yang-etal-2019-enhancing}. For example, for the entities in the given context of Table \ref{intro}, many potentially related concepts~(e.g., \texttt{run over}, \texttt{cross street})
can be inferred and predicted based on external commonsense knowledge bases such as ConceptNet~\cite{Speer2012Representing} and ATOMIC~\cite{sap2019atomic}. These knowledge bases contain abundant semantic knowledge of concepts and inferential knowledge for commonsense reasoning. We enhance GPT-2 with such knowledge 
by post-training the model on the knowledge examples constructed from these knowledge bases, which can provide additional crucial information for story generation. 
Empirical experiments demonstrate that training with millions of such examples helps improve the coherence and logicality of generated stories.} Meanwhile, we adopt multi-task learning to address the problem of handling causal and temporal dependencies. We combine the generation objective with an auxiliary multi-label classification objective, which requires distinguishing true stories from fake stories that are constructed by randomly shuffling the sentences, replacing a sentence with a negatively sampled one, or repeating a sentence in an original story. 
The additional classification task empowers our model to better capture the logicality in a story implicitly, namely, modeling the causal and temporal dependencies, and inter-sentence coherence, and avoiding repetition.  



The main contributions of this paper are summarized as follows:

\begin{itemize}
    \item We propose a knowledge-enhanced pretraining model for commonsense story generation by extending 
    GPT-2 with external commonsense knowledge. The model is post-trained on the knowledge examples constructed from ConceptNet and ATOMIC, thereby improving long-range coherence of generated stories.
    
    \item To generate reasonable stories, we adopt a classification task to distinguish true stories from auto-constructed fake stories. The auxiliary task makes the model implicitly capture the causal, temporal dependencies between sentences and inter-sentence coherence, and lead to less repetition.
    
    \item We conduct extensive experiments with automatic and manual evaluation. Results show that our model can generate more reasonable stories than strong baselines, particularly in terms of logicality and global coherence.\footnote{{Our implementation is available at \url{https://github.com/thu-coai/CommonsenseStoryGen}}, and demo is available at \url{http://coai.cs.tsinghua.edu.cn/static/CommonsenseStoryGen}.}

\end{itemize}

\section{Related Work}
\noindent\textbf{Neural Story Generation} \\
Many existing neural story generation models generated stories by conditioning upon various contents such as images~\cite{huang2016visual} and short text descriptions~\cite{Jain2017Story}. Different from these studies, we consider the setting of open-ended story generation from only a limited leading context in this paper. For this task, prior studies have attempted to build specific sentence representations by modeling story entities and events to simplify the dependencies between sentences~\cite{Ji2017Dynamic,Elizabeth2018Neural}. Another line is to decompose story generation into separate steps~\cite{Lara2018Event,fan2018hierarchical,wang2016chinese,xu2018skeleton,yao2018plan,fan-etal-2019-strategies}. These models usually focused on first planning story sketches and then generating sentences from the sketches.
However, improving pretrained models to generate commonsense stories is yet to be well investigated.

\noindent\textbf{Pretraining} \\
Recently large-scale pretraining models have been widely developed in various NLP tasks. Some work leveraged pretraining to provide better language representations in word level~\cite{mikolov2013efficient,pennington2014glove,peters2018deep} or sentence level~\cite{le2014distributed,NIPS2015_5950} for various downstream task-specific architectures. However, \citeauthor{radford2018improving}~(\citeyear{radford2018improving}) and \citeauthor{Devlin2018BERT}~(\citeyear{Devlin2018BERT}) suggests these complex task-specific architectures are no longer necessary, and it is sufficient to merely fine-tune pretrained task-independent transformer language models for downstream tasks. \citeauthor{mehri-etal-2019-pretraining}~(\citeyear{mehri-etal-2019-pretraining}) explored different pretraining methods based on language models for dialogue context representation learning. Furthermore, \citeauthor{radford2019language}~(\citeyear{radford2019language}) demonstrate pretrained language models~(i.e., GPT-2) can perform downstream tasks better than state-of-the-art models even in zero-shot setting~(i.e., without any fine-tuning on task-specific data).  \citeauthor{wolf2019transfertransfo}~(\citeyear{wolf2019transfertransfo}) fine-tuned GPT-2 for personalized conversation generation, which obtains very competitive results in the challenge. {However, as previous studies~\cite{see2019massively,holtzman2019curious} observed, transferring GPT-2 directly to open-ended text generation still suffers from several issues such as repetition or lack of knowledge and inter-sentence coherence with different decoding algorithms.} 
Besides, although \citeauthor{song2019mass}~(\citeyear{song2019mass}) and \citeauthor{dong2019unified}~(\citeyear{dong2019unified}) extended the language model to support encoder-decoder framework~\cite{sutskever2014sequence}, we build our model based on GPT-2 due to its simplicity and broad applicability. 


\noindent\textbf{Commonsense Knowledge} \\
Incorporating commonsense knowledge is necessary and beneficial for language inference~\cite{Peter2011Types,Samuel2015Large, Hannah2018event2mind}, reading comprehension~\cite{Todor2018kr,Rashkin2018Modeling}, and particularly for open-ended language generation, which usually requires external knowledge to enrich the limited source information. Commonsense knowledge has been demonstrated to significantly improve dialogue generation ~\cite{Zhou2018Commonsense}, story ending generation~\cite{guan2019story}, and essay generation from given topics~\cite{yang-etal-2019-enhancing}. 
And recently, some work also attempted to integrate external commonsense knowledge into pretrained models such as BERT~\cite{Devlin2018BERT} to enhance language representation for reading comprehension~\cite{yang-etal-2019-enhancing-pre} and other knowledge-driven NLP tasks like entity typing and relation classification~\cite{zhang2019ernie}. Besides, \citeauthor{sun2019ernie}~(\citeyear{sun2019ernie}) improved BERT on Chinese NLP tasks by multi-stage knowledge masking strategy to integrate phrase and entity level knowledge into the language representation. {Moreover, \citeauthor{bosselut2019comet}~(\citeyear{bosselut2019comet}) transferred the implicit knowledge from GPT-2 by fine-tuning the model to generate an object given the subject and a relation as input in commonsense knowledge graphs, i.e., automatic knowledge base construction. However, the low novelty of the generated objects showed that it could still be difficult for GPT-2 to generate commonsense texts solely based on its implicit knowledge. Therefore, we target at integrating external knowledge into GPT-2 for generating more reasonable commonsense stories.}


\noindent\textbf{Multi-Task Learning} \\
Incorporating other auxiliary task objectives to complement the primary goal has been shown to improve the performance in many NLP tasks such as sentiment classification~\cite{yu2016learning} and conversation generation~\cite{zhao2017learning}. 
Recently, multi-task learning was also used to pretrain language models to capture dependencies in context~\cite{Devlin2018BERT,mehri-etal-2019-pretraining} and further improve pretrained models' representation power during fine-tuning~\cite{wolf2019transfertransfo}.


\begin{table*}[!ht]
\scriptsize
\centering
\begin{tabular}{l l l}
    \toprule
    \textsf{\textbf{Knowledge}} &\multirow{2}{*}{\textsf{\textbf{Original Triples}}} & \multirow{2}{*}{\textsf{\textbf{Examples of Transformed Sentences}}}\\
    \textsf{\textbf{Bases}} & & \\
    \midrule
    \multirow{2}{*}{\textsf{ConceptNet}}&\textsf{(eiffel tower, \textbf{AtLocation}, paris)} & \textsf{eiffel tower \textbf{is at} paris.}\\
    &\textsf{(telephone, \textbf{UsedFor}, communication)} & \textsf{telephone \textbf{is used for} communication.}\\
    \midrule
    \multirow{2}{*}{\textsf{ATOMIC}}&\textsf{(PersonX dates for years, \textbf{oEffect}, continue dating)} & \textsf{PersonX dates for years. \textbf{PersonY will} continue dating.}\\
    &\textsf{(PersonX cooks spaghetti, \textbf{xIntent}, to eat)} & \textsf{PersonX cooks spaghetti. \textbf{PersonX wants} to eat.}\\
    \bottomrule
\end{tabular}
\caption{Examples of template-based transformation of triples in knowledge bases. Phrases in \textbf{bold} represent the original and transformed relations.}
\label{kg_transform}
\end{table*}

\section{Methodology}
\begin{figure}[!ht]
\centering
\includegraphics[width=3in]{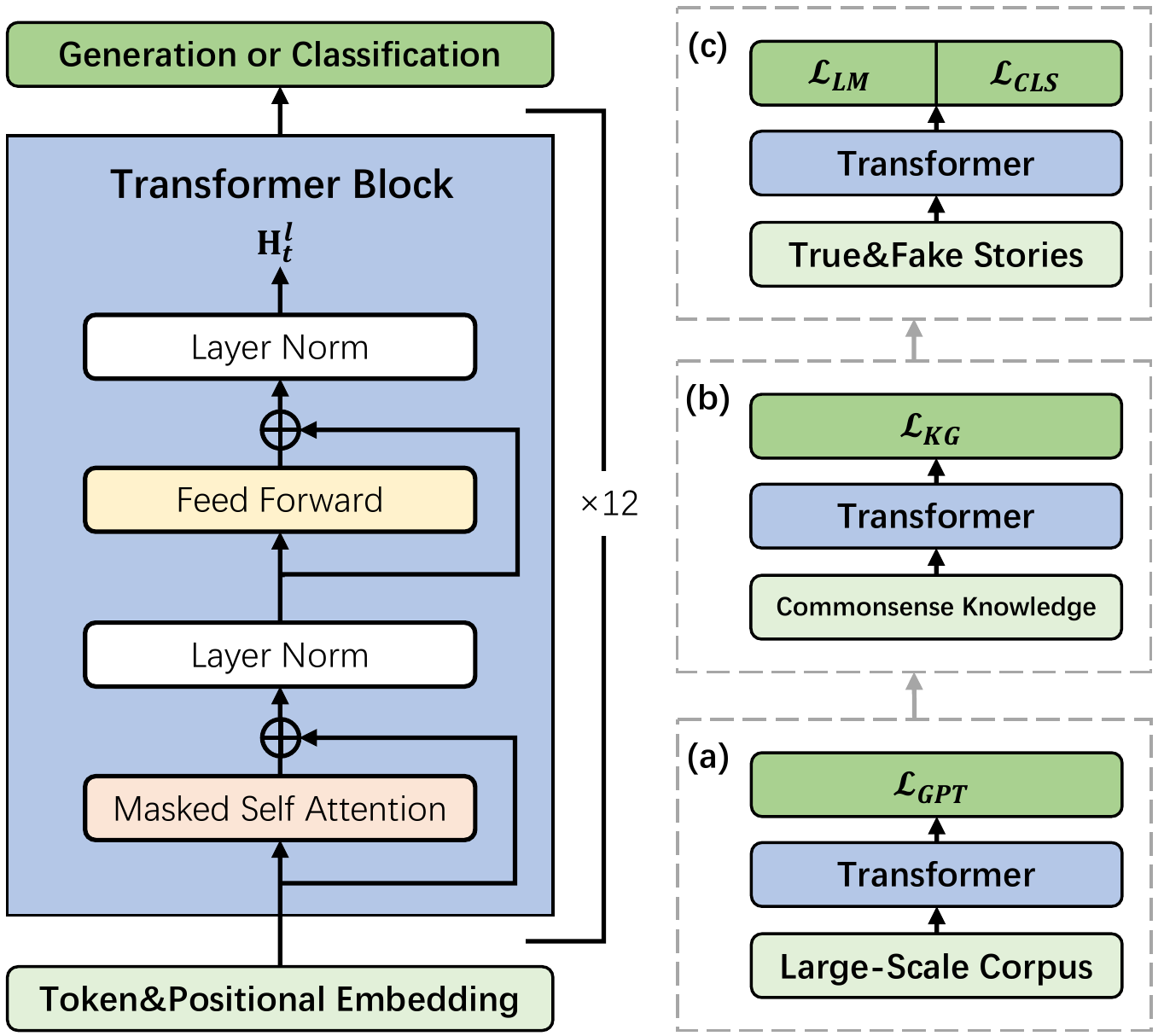}
\caption{Transformer block architecture~(left) and training framework~(right). 
We divide the whole training framework into the following three stages. Train the language model (a) with a large-scale corpus, in which stage we directly inherit the pretrained model parameters from~\citeauthor{radford2019language}~(\citeyear{radford2019language}), (b) with commonsense knowledge from external knowledge bases, and (c) with true and auto-constructed fake stories by multi-task learning for story generation and classification. $\mathcal{L}_{GPT}$, $\mathcal{L}_{KG}$, $\mathcal{L}_{LM}$ and $\mathcal{L}_{CLS}$ are the corresponding loss functions in different stages respectively. }

\label{model}
\end{figure}

    The task in this work can be defined as follows: given a one-sentence story beginning $X$ as the leading context, the model should continue to complete a $K$-sentence story $Y$ with a reasonable plot. The sentences in a generated story should have reasonable logical connections, causal relationships, and temporal dependencies with each other and with the given beginning. To this end, we devise a novel framework to leverage knowledge and handle the causal and temporal dependencies, as Figure \ref{model} shows. 
    

\subsection{Pretrained Transformer Language Model}


The transformer architecture is a general model used in language modeling~\cite{vaswani2017attention}, which consists of multiple transformer blocks of multi-head self-attention followed by layer-normalization and fully-connected layers.  \citeauthor{radford2019language}~(\citeyear{radford2019language}) used a 12-layer decoder-only transformer (GPT-2), i.e., a left-to-right language model, with masked self-attention heads which are constrained in that every token can only attend to its left context. Formally, the objective in this stage is to minimize the following negative likelihood: 
\begin{align}
    \mathcal{L}_{GPT}&=-\sum_{t=1}^{|u|}{\rm log}P(u_t|u_{<t}),\\
    P(u_t|u_{<t}) &= \textbf{softmax}(\textbf{H}_{t}^{L}\textbf{W}+\textbf{b}),\\
    \textbf{H}_t^{l} &= \texttt{block}(\textbf{H}_{<t}^{l-1}), l\in [1, L],\\
    \textbf{H}_t^0&=E_t+P_t,
\end{align}
where $u$ is an utterance with $|u|$ tokens in total from the training corpus, $u_t$ is the $t$-th tokens in $u$, $\textbf{H}^{l}_t$ is the $l$-th layer's output at the $t$-th position computed through the transformer block with the masked self attention mechanism, and $\textbf{H}^0_t$ is a summation of token embedding $E_t$ and positional embedding $P_t$ for the $t$-th token.  

GPT-2 network is pretrained on a large-scale corpus 
but still suffers from many issues such as lack of necessary knowledge for commonsense story generation as aforementioned. 
Therefore, in this work we improve GPT-2 for generating more reasonable stories with external commonsense knowledge. 


\subsection{Training with Commonsense Knowledge}


Commonsense  knowledge  can  facilitate  language  comprehension and generation, as reported in a notable work for dialog generation~\cite{Zhou2018Commonsense}. To leverage commonsense knowledge in pretrained language models, we resort to existing large-scale knowledge bases ConceptNet~\cite{li-16} and ATOMIC~\cite{sap2019atomic}.

The ConceptNet dataset\footnote{\url{http://www.conceptnet.io/}} consists of triples obtained from the Open Mind Common Sense entries in ConceptNet 5~\cite{Speer2012Representing}. It contains 34 relations in total and represents each knowledge triple by $R=(h,r,t)$, meaning that head concept $h$ has the relation $r$ with tail concept $t$, e.g., \texttt{(cross street, Causes, accident)}. And the ATOMIC dataset\footnote{\url{https://homes.cs.washington.edu/~msap/atomic/}} is an atlas of everyday commonsense reasoning, containing a mass of textual description of inferential knowledge organized as typed \textit{if-then} triples.
For example, a typical \textit{if-then} triple is \texttt{(PersonX pays PersonY a compliment, xIntent, to be nice)}, where \texttt{xIntent} is the relation between the head and tail events standing for If-Event-Then-Mental-State.

We implicitly introduce the knowledge to the pretrained language model by post-training on knowledge-augmented data. Some work has attempted to explicitly incorporate commonsense knowledge into language generation~\cite{Zhou2018Commonsense,guan2019story,yang-etal-2019-enhancing}. However, all these works assume there is an alignment between the training data and the knowledge bases. Therefore, they suffer from the following issues: (1) It is difficult to match the events extracted from the training data with those stored in KB. 
(2) Learning and utilizing multi-hop triples in knowledge graphs is costly in time due to the large-scale size. (3) Most of KB triples do not appear in the task-specific training data, so that those absent triples are not fully utilized in existing models.
Fortunately, our model is trained on the knowledge bases directly, which can effectively ease the above limitations. 

We transform the commonsense triples in ConceptNet and ATOMIC into readable natural language sentences using a template-based method~\cite{Levy2017Zero}, as illustrated in Table~\ref{kg_transform}. We do not use roughly concatenated triples in order to avoid introducing additional special tokens~(e.g., \texttt{UsedFor} in ConceptNet and \texttt{oEffect} in ATOMIC), or break the syntactic features contained in the pretrained language model~\cite{alt-etal-2019-fine}, which are essential for following story generation.
And then the language model is post-trained on the transformed sentences to learn commonsense knowledge between entities and events by minimizing the negative likelihood of predicting the next token:
\begin{align}
    \mathcal{L}_{KG}=-\sum_{t=1}^{|r|}{\rm log} P(r_t|{r}_{<t}),
\end{align}
where $r$ is a transformed sentence with $|r|$ tokens in total, and ${r}_t$ is the $t$-th token in $r$. In this way, we can incorporate commonsense knowledge into GPT-2 implicitly.


\subsection{Multi-Task Learning}

\begin{figure}[!ht]
\centering
\includegraphics[width=3in]{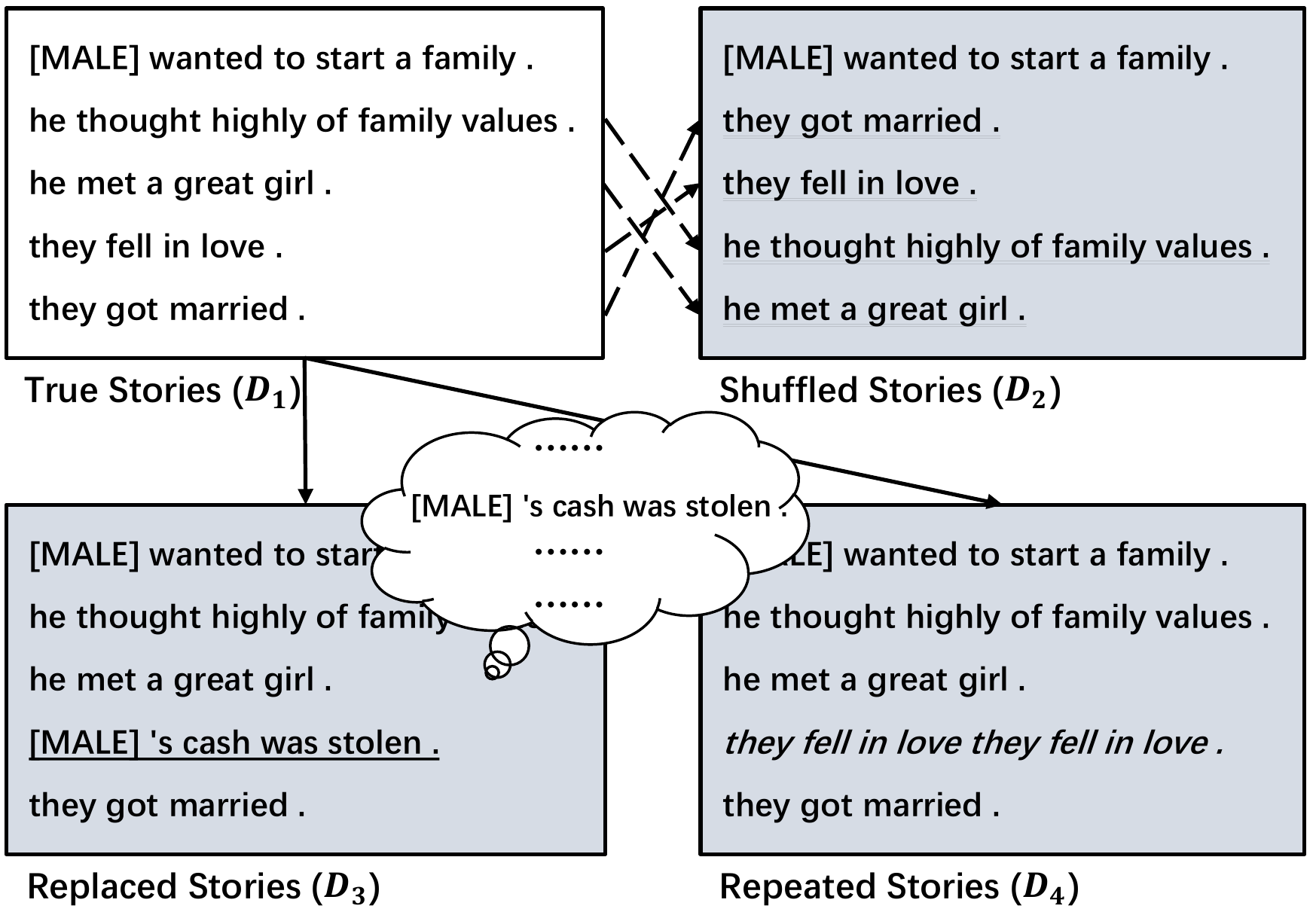}
\caption{An example of fake story construction. The shuffled sentences are indicated by dashed lines, the replaced sentence is \underline{underlined}, and the repeated one is in \textit{italic}.}
\label{data_gen}
\end{figure}

In order to encourage our model to generate reasonable stories in logic, we add an auxiliary classification task to the generation task during fine-tuning on the ROCStories corpus. The task requires distinguishing true stories from fake stories. We first construct three additional sets of fake stories by shuffling the sentences, replacing a sentence with a negatively sampled one, and randomly repeating a sentence in an original story. 
Notably, the above operations are performed only on the following $K$ sentences of a story, i.e., not including the leading context (the beginning). For simplicity, we denote the true story set and three manually constructed fake story sets with $D_1,D_2,D_3$ and $D_4$ respectively, as illustrated in Figure \ref{data_gen}.

{Our main finding is that training a language model to distinguish the reasonable stories from those with disordered logic, unrelated topics, or repeated plots is helpful to generate more reasonable stories in terms of logic and coherence.} 
We add an additional classification layer at the last layer of the transformer language model in a multi-task setting. The classifier takes as input the hidden states of the last transformer block and computes a score through a softmax layer over $D_1$, $D_2$, $D_3$, and $D_4$, formally as follows:
\begin{align}
    P(l_s|s) = \textbf{softmax}(\frac{1}{|s|}\sum_{t=1}^{|s|} \textbf{H}^L_t\textbf{W}_L + \textbf{b}_L),
\end{align}
where $s$ is a true or fake story and contains $|s|$ tokens, $\textbf{H}^L_t$ is the hidden state of the $L$-th block layer~(i.e., the last layer) of the transformer language model when encoding the story, $l_s$ is predicted to indicate which dataset~($D_i$) the story~($s$) belongs to, and $\textbf{W}_L$ and $\textbf{b}_L$ are the trainable parameters of the additional classifier.

\begin{figure}[!ht]
\centering
\includegraphics[width=2.8in]{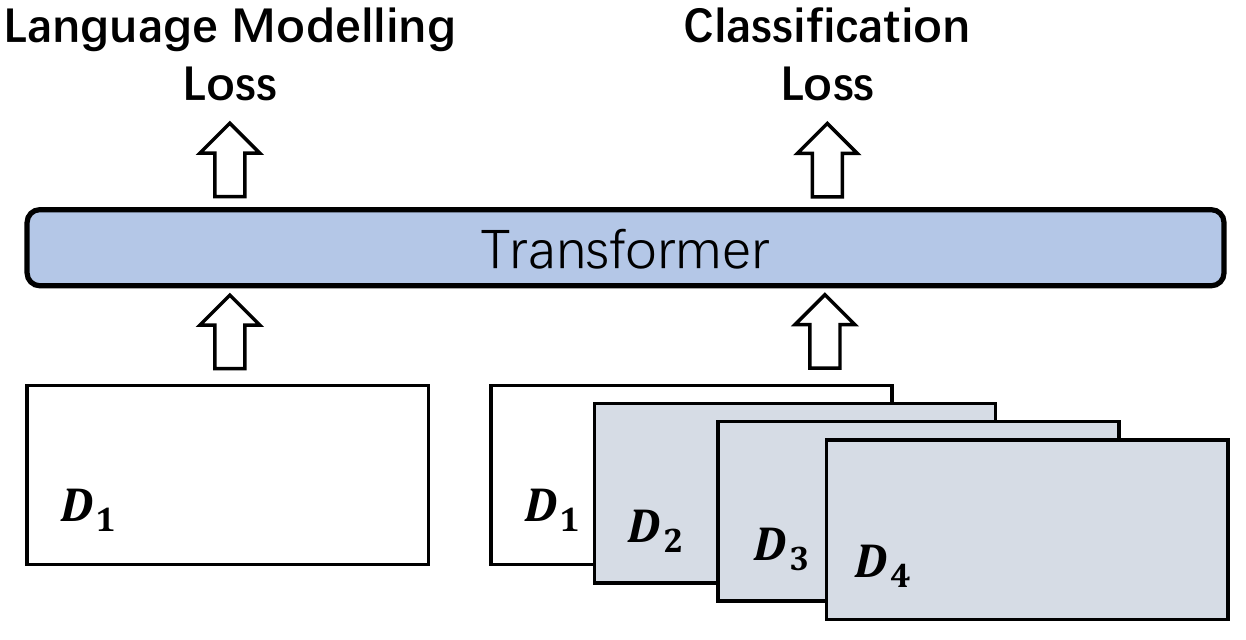}
\caption{Multi-task learning diagram. $D_1$ is the true story dataset, while $D_2,D_3$ and $D_4$ are the auto-constructed fake stories transformed from $D_1$. Note that the language modeling loss is optimized only on the true stories, but the classification loss on both true and fake ones.}
\label{multi_task}
\end{figure}

As illustrated in Figure \ref{multi_task}, the loss function $\mathcal{L}_{ST}$ of the full model is computed as follows:
\begin{align}
    &\mathcal{L}_{ST} = \mathcal{L}_{LM} + \lambda\mathcal{L}_{CLS},\\
    &\mathcal{L}_{LM} = -\sum_{t=1}^{|s|}{\rm log} P(s_t|s_{<t}),s\in D_1,\\
    &\mathcal{L}_{CLS} = -{\rm log} P(l_s=\Tilde{l}_s|s), s\in D_1,D_2,D_3,D_4,
\end{align}
where $s$ is a story containing $|s|$ tokens, ${s}_t$ is the $t$-th token of $s$, $\mathcal{L}_{LM}$ is the language modeling loss, $\mathcal{L}_{CLS}$ is the classification loss, and $\Tilde{l}_s$ indicates the correct $D_i$ which the story $s$ is sampled from. 
$\lambda$ is an adjustable scale factor. 

\section{Experiments}
\subsection{Dataset}
    We evaluated our model on the ROCStories corpus \cite{Mostafazadeh2016A}. The corpus contains 98,162 five-sentence stories for evaluating story understanding. 
    The original task is designed to select a correct story ending from two candidates~\cite{zhou2019story}, while our task is to generate a reasonable story given the first sentence of a story ~(i.e., $K$, namely the number of generated sentences, is four in our setting). Following~\citeauthor{radford2019language}~(\citeyear{radford2019language}), the stories are tokenized using byte pair encoding (BPE) with a vocabulary of 50,257 items. The average number of tokens in $X/Y$~(i.e., the beginning/the following $K$ sentences in a story)  
    is 13.39/50.00 with BPE, while the model uses pretrained positional embeddings with a maximal sequence length of 1024 tokens. 
    
    As for the knowledge bases, we used the 605k version of ConceptNet. The second KB we used contains 709k records from the 877k tuples of ATOMIC after transformation and deduplication. We randomly selected stories and knowledge sentences for training/validation/test respectively, as shown in Table \ref{dataset}. Since the ROCStories dataset is rather small for generation, we made delexilization by replacing all the names in stories with special placeholders ``[MALE]$"$, ``[FEMALE]$"$, and ``[NEUTRAL]$"$ for male, female and unknown names respectively. Besides, ``PersonX$"$ and ``PersonY$"$ in ATOMIC are replaced by ``[MALE]$"$ and ``[FEMALE]$"$ as well.
  \begin{table}[!ht]
\footnotesize
\centering
\begin{tabular}{cccc}
    \toprule
    \textbf{Dataset} & \textbf{Training} & \textbf{Validation} & \textbf{Test}\\
    \midrule
    \textbf{ROCStories} & 88,344 & 4,908 & 4,909 \\
    \textbf{ConceptNet} & 600,000 & 2,400 & 2,400\\
    \textbf{ATOMIC} & 574,267 & 70,683 & 64,456\\
    \bottomrule
\end{tabular}
\caption{Statistics of datasets and knowledge bases.}
\label{dataset}
\end{table}  
\subsection{Baselines}

\begin{table*}[!ht]
\scriptsize
\centering
\begin{tabular}{lcccccccc}
    \toprule
    \textbf{Models} & \textbf{PPL} & \textbf{BLEU-1} & \textbf{BLEU-2} & \textbf{Coverage} & \textbf{Repetition-4(\%)} & \textbf{Distinct-4}(\%)\\
    \midrule
    \textbf{ConvS2S} & N/A  & 0.312 & 0.132&13.64  &22.87 &72.78\\
    \textbf{Fusion} & N/A & 0.322 & 0.137&12.02 & 24.23& 72.82\\
    \textbf{Plan\&Write} & N/A &0.308 & 0.126& 13.38 & 17.06&67.20\\
    \textbf{SKRL} & N/A & 0.267 &0.088 &10.82 &18.34 & 69.42\\
    \textbf{DSRL} & N/A & 0.293&0.117 &10.38 &\textbf{15.36}& 73.08\\
    \midrule
    \textbf{GPT-2 (Scratch)}&11.82 & 0.311 & 0.134&10.76 &22.87& 73.33\\
    \textbf{GPT-2 (Pretrain)} & 33.50 &0.257 & 0.085&8.04 &39.22&64.99 \\
    \textbf{GPT-2 (Fine-tune)}  & 7.96 & 0.322 & 0.141 &12.40 & 29.41 &73.85\\
    \midrule
    \textbf{Ours}  & 7.85 &\textbf{0.326} &\textbf{0.143} &\textbf{18.48} &{21.93}&78.96 \\
    ~~\textbf{w/o~Pretrain} & 11.04 &0.316 &0.134&16.33 &21.52&77.17\\
    ~~\textbf{w/o~Knowledge}  & \textbf{7.70} &0.314&0.136 &13.95 &25.08&73.24\\
    ~~\textbf{w/o~Multi-task} & 8.04 & 0.324 & 0.140&17.19&24.40 &\textbf{79.43}\\
    \midrule
   \textit{\textbf{Golden Story}}& \textit{N/A} & \textit{N/A} & \textit{N/A} &\textit{19.28}&\textit{7.64}&\textit{89.51}\\
    \bottomrule
\end{tabular}
\caption{Automatic evaluation results. The best performance is highlighted in \textbf{bold}. And the results of golden story are in \textit{italic}. The perplexity scores marked with N/A are not comparable with ours because the corresponding models tokenize stories by words rather than by byte pair encodings used in GPT-2.}
\label{auto-eva}
\end{table*}

We compared our models with the following state-of-the-art baselines:

\noindent\textbf{Convolutional Seq2Seq~(ConvS2S):} It directly generates a story conditioned upon the beginning based on a convolutional seq2seq model~\cite{Gehring2017Convolutional} with decoder self-attention.

\noindent\textbf{Fusion Convolutional Seq2Seq Model~(Fusion):} It generates a story by first pretraining a convolutional seq2seq model, and then fixing the model and providing it to the second clone model with fusion mechanism~\cite{fan2018hierarchical}.

\noindent\textbf{Plan\&Write:} It first generates a sequence of keywords as planning, conditioned upon the input; and then generates a story based on the planned keywords \cite{yao2018plan}. 
During training, one keyword is extracted from each sentence with RAKE algorithm~\cite{Rose2010Automatic}.

\noindent\textbf{Skeleton-based Model with Reinforcement Learning~(SKRL):} The model first generates a compressed story including the most critical phrases, called skeleton, and then generates a story conditioned upon the skeleton. The skeleton is automatically learned by reinforcement learning~\cite{xu2018skeleton}.

\noindent\textbf{Decomposed Model with Semantic Role Labeling (DSRL):} It first generates a predicate-argument structure conditioned upon the beginning and then generates a story by surface realization on top of the structure. The structures are identified by semantic role labelling~\cite{fan-etal-2019-strategies}.

We also made comparisons with GPT-2 in different settings as follows:

\noindent\textbf{GPT-2~(Scratch):} The network architecture is the same as GPT-2, but the model is only trained on ROCStories without any pretrained parameters.

\noindent\textbf{GPT-2~(Pretrain):} This model directly used the public checkpoint of pretrained parameters\footnote{The pretrained model is available at \url{https://github.com/openai/gpt-2}.} 
for story generation. Following \citeauthor{radford2019language}~(\citeyear{radford2019language}), stories are generated in a zero-shot setting. To induce story generation behavior, we conditioned the language model on a context of example stories, and then sample sentences from the model after a final prompt of story beginning. We used the first $K$ generated sentences as the generated story.

\noindent\textbf{GPT-2~(Fine-tuning):} This model is fine-tuned on the ROCStories corpus from the public checkpoint of pretrained parameters.


Furthermore, we also conducted ablation tests by removing the proposed components respectively to investigate the influence of each component with the same network structure.






\subsection{Experiment Settings}
We set the parameters by following the small version of~\citeauthor{radford2019language}~(\citeyear{radford2019language})'s design: the language model is equipped with 12 layers, 768-dimensional hidden states, and 12 attention heads. The batch size is 10 during training on the ROCStories corpus using Adam optimizer with an initial learning rate of 1e-4. The scale factor $\lambda$ is set to 0.05. 
And we generated stories using a top-$k$ sampling scheme~\cite{fan2018hierarchical} with $k$=40 and a softmax temperature of 0.7~\cite{goodfellow2016deep} to balance the trade-off between diversity and fluency. We applied these settings to all the baselines.


\subsection{Automatic Evaluation}

\noindent\textbf{Evaluation Metrics} 
We adopted the following automatic metrics to evaluate the generation performance in the entire test set. (1)~\textbf{Perplexity~(PPL)}. Smaller perplexity scores indicate better fluency in general. 
(2)~\textbf{BLEU}. BLEU~\cite{Papineni2002IBM} evaluates $n$-gram overlap between a generated story and a human-written story. However, BLEU is usually inappropriate for open-ended text generation~\cite{fan2018hierarchical} since there are multiple plausible stories for the same input but only one story is given in the dataset. And BLEU scores will become extremely low for large $n$. We thus experimented with $n$=1,2. (3)~\textbf{Coverage}. To access the effect of incorporating commonsense knowledge, we calculated the coverage score as the average number of commonsense triples matched in each generated story, which requires both head and tail entities/events appears in the same story. (4) \textbf{Repetition}. We measured the redundancy of stories by computing repetition-4, the percentage of generated stories that repeat at least one 4-gram~\cite{Shao2019Planning}. (5) \textbf{Distinct}. To measure the generation diversity, we adopted distinct-4~\cite{li2015diversity}, the ratio of distinct 4-grams to all the generated 4-grams.


\begin{table*}[!ht]
\scriptsize
\centering
\begin{tabular}{lllllllll}
    \toprule
    \multicolumn{1}{c}{\multirow{2}{*}{\textbf{Models}}} & \multicolumn{4}{c}{~~~~~~\textbf{Grammaticality}} & \multicolumn{4}{c}{~~~~~~\textbf{Logicality}}\\
    &~~~~~~\textbf{Win~(\%)} & \textbf{Lose~(\%)} & \textbf{Tie~(\%)} & $\kappa$ & ~~~~~~\textbf{Win~(\%)} & \textbf{Lose~(\%)} & \textbf{Tie~(\%)} & $\kappa$\\
    \midrule
    \textbf{Ours vs. Fusion} &~~~~~~50.0** &27.0 &23.0&0.421&~~~~~~57.0**& 28.0& 15.0&0.455\\
    \textbf{Ours vs. DSRL} &~~~~~~58.0** &24.0&18.0 & 0.441&~~~~~~58.0**&29.0&12.0 &0.475\\
    \midrule
    \textbf{Ours vs. GPT-2 (Scratch)} &~~~~~~54.0**& 24.5 &21.5&0.385 &~~~~~~54.0** &26.0 &20.0&0.304\\
    \textbf{Ours vs. GPT-2 (Pretrain)} &~~~~~~52.0** &31.5 &16.5&0.483 &~~~~~~56.5** &32.5 &11.0&0.493\\
    \textbf{Ours vs. GPT-2 (Fine-tune)} &~~~~~~42.0**& 28.0&30.0& 0.344&~~~~~~51.0**&27.5 & 21.5 &0.371\\
    \midrule
    \textbf{Ours vs. Ours w/o Pretrain}&~~~~~~51.0** & 31.0&18.0&0.378 &~~~~~~56.0** &28.0 &16.0 & 0.375\\
    \textbf{Ours vs. Ours w/o Knowledge} &~~~~~~46.0** & 23.0& 21.0&0.289&~~~~~~48.0** &29.0 &23.0&0.314\\
    \textbf{Ours vs. Ours w/o Multi-task} & ~~~~~~37.5 & 31.0 &31.5 &0.313&~~~~~~48.5** &25.5 &26.0&0.297\\
    \bottomrule
\end{tabular}
\caption{Manual evaluation results. The scores indicate the percentages of \textit{Win}, \textit{Lose} or \textit{Tie} when our model is compared with a baseline. $\kappa$ denotes Fleiss' kappa (all are \textit{fair agreement} or \textit{moderate agreement}). The scores marked with * mean p-value< 0.05 and ** indicates p-value< 0.01 in sign test.}
\label{manual-eva}
\end{table*}

\noindent\textbf{Results}
The results of automatic evaluation are shown in Table \ref{auto-eva}. Note that the perplexity scores of some baselines are not comparable with ours because they tokenize stories by words rather than by byte pair encodings as used in GPT-2. Thus, we did not provide these scores. Our model outperforms the variants of GPT-2 in terms of perplexity, and has higher BLEU scores than all the baselines, indicating better fluency and more overlaps with the reference stories. Our model also has higher knowledge coverage 
and distinct-4 scores, showing that our model can generate more diverse stories with more abundant knowledge.
However, we observed that pretraining might lead to more severe repetition by comparing three variants of GPT-2. Our model effectively improves the situation but still performs worse than the baselines with task-specific architectures, for instance, the planning-based models~(e.g., DSRL). 
Fortunately, \citeauthor{see2019massively}~(\citeyear{see2019massively}) showed that increasing $k$ for top-$k$ sampling could alleviate the repetition issue. Besides, compared with training from scratch, fine-tuned GPT-2 performs much better in fluency (lower perplexity scores) but suffers from worse repetition, and only improve slightly in coverage and diversity. Furthermore, pretrained GPT-2 has the lowest coverage and distinct-4, which further verifies our hypothesis that GPT-2 lacks the necessary knowledge to expand a story plot. 

As for the ablation test, our model without pretraining has significantly higher perplexity, 
indicating that pretraining contributes to story fluency. {When removing external knowledge, coverage and distinct-4 drop while repetition-4 rises substantially, suggesting that post-training on millions of knowledge sentences can effectively enhance the language model's ability to generate stories with more commonsense knowledge, although we do not explicitly utilize knowledge during fine-tuning on ROCStories.} Besides, removing multi-task learning leads to slightly better distinct-4 but causes much higher repetition-4, indicating that the classification loss is of great help for reducing redundancy.

{We also provide the performance of our model on the auxiliary story classification task and the predicted proportional distribution of the generated stories by different models on the four story types with the auxiliary story classifier, as shown in Table~\ref{cls_acc}. Both metrics are computed on 1,000 samples from the test set. We can observe that it is relatively easier to detect fake stories with repeated plots~($D_4$) than those with disordered logic~($D_2$) and unrelated topics~($D_3$). When using the auxiliary story classifier to classify the generated stories, pretrained GPT-2 is considered to generate more fake stories, with only 15.83\% stories of type $D_1$, which agrees with the previous automatic evaluation especially in terms of repetition. Besides, our model performs better than baselines, indicating that the external knowledge and the auxiliary task can encourage our model to generate more reasonable stories. }

\begin{table}[!ht]
\footnotesize
\centering
\begin{tabular}{lcccc}
\toprule
\textbf{Story types}& $\mathbf{D_1}$ & $\mathbf{D_2}$ & $\mathbf{D_3}$ & $\mathbf{D_4}$\\
\midrule
\textbf{{F1 score}} & {0.80} & {0.81} & {0.88}& {0.98} \\
\midrule
\midrule
{\textbf{Models}} &\multicolumn{4}{c}{\textbf{Proportional Distribution~(\%)}} \\
\midrule
\textbf{GPT-2~(Pretrain)} &15.83 &40.8 &39.36 &4.01 \\
\textbf{GPT-2~(Fine-tune)} &86.94&9.98&2.93&0.15 \\
\midrule
\textbf{Ours} &90.12 &7.98 &1.86 &0.04 \\
~~\textbf{w/o Knowledge} & 87.76& 9.51& 2.67&0.06 \\
~~\textbf{w/o Multi-task} & 88.69& 9.07& 2.02&0.22 \\

\bottomrule
\end{tabular}
\caption{Final prediction {\textbf{F1 score}} of our model on the auxiliary story classification task in terms of the four types of story sets respectively, and the \textbf{proportional distribution} of the predicted story types of the generated stories by different models.}
\label{cls_acc}
\end{table}


\begin{table}[!ht]
\footnotesize
\centering
\begin{tabular}{lcc}
\toprule
\textbf{Models} & \textbf{BR~(\%)} & \textbf{LR~(\%)}\\
\midrule
\textbf{GPT-2~(Pretrain)} &59.3 & 44.8\\
\textbf{GPT-2~(Fine-tune)}& 73.4 & 69.6\\
\midrule
\textbf{Ours} &76.2 & 72.7\\
~~\textbf{w/o Knowledge}& 74.9& 71.5\\
~~\textbf{w/o Multi-task} &75.7 & 70.4\\
\bottomrule
\end{tabular}
\caption{Accuracy of beginning ranking and logic ranking. 
Larger scores are better.
}
\label{brlr}
\end{table}

{Following \citeauthor{fan2018hierarchical}~(\citeyear{fan2018hierarchical}) and \citeauthor{see2019massively}~(\citeyear{see2019massively}), we computed \textit{beginning ranking accuracy}~(BR) to measure how strongly the output of a model is coherent with the beginning, and \textit{logic ranking accuracy}~(LR)  to measure the ability of capturing the causal and temporal dependencies in the context. 
For BR, we first sampled 9 negative beginnings (first sentence) for a true story, and then calculated the perplexity of the 10 stories. If the true story has the lowest perplexity by our model, it is regarded as a correct prediction. 
As for LR, since each story in ROCStories consists of five sentences, we produced four shuffled versions by switching each pair of adjacent sentences. We then used our model to score the five stories with perplexity. A prediction is regarded as correct if the true story has the lowest score. 
We randomly sampled 1,000 human-written stories from the test set in our evaluation. As shown in Table~\ref{brlr}, the external knowledge and multi-task learning effectively promote the coherence and help capture inter-sentence dependencies in the context.} 

\subsection{Manual Evaluation}
To evaluate the fluency and logic of generated stories, we conducted pair-wise comparisons with two strong baseline models~(Fusion and DSRL) that performed best in automatic evaluation, three variants of GPT-2, and three ablated models of ours.  For manual evaluation, we randomly sampled 200 stories from the test set and obtained 1,800 stories from the nine models. For each pair of stories (one by our model and the other by a baseline, along with the beginning), three annotators were hired to give a preference (win, lose, or tie) in terms of two metrics respectively. 
We resorted to a crowdsourcing service Amazon Mechanical Turk (AMT) for annotation, and we adopted majority voting to make final decisions among the three annotators.

\begin{figure}[!ht]
\centering
\includegraphics[width=2.5in]{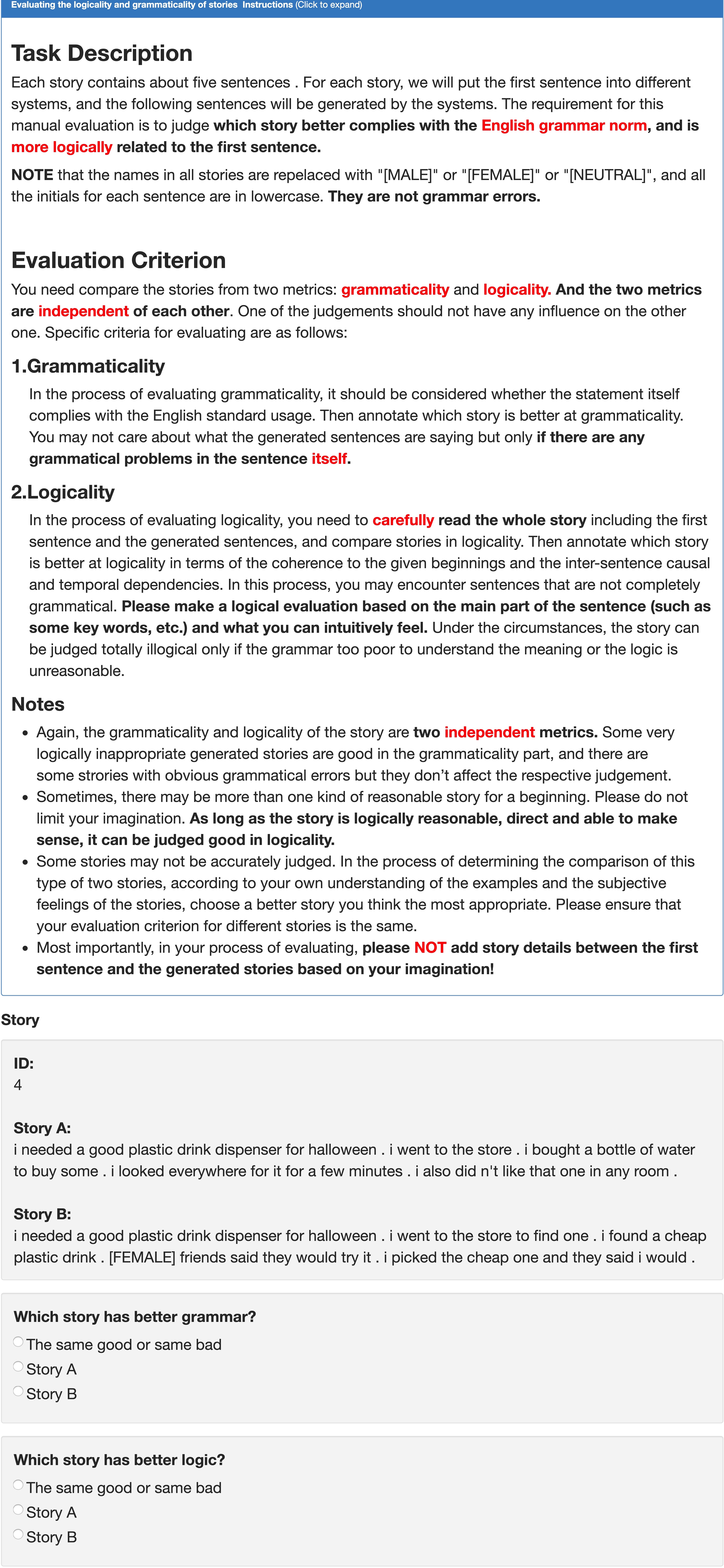}
\caption{A screenshot of the annotation on AMT for manual evaluation.}
\label{AMT}
\end{figure}

\noindent\textbf{Evaluation Metrics} {We evaluated the models from the following two perspectives:
\textbf{grammaticality} to indicate whether a story is natural and fluent, and \textbf{logicality} to indicate whether a story is coherent to the given beginning and reasonable in terms of causal and temporal dependencies in the context. 
Note that the two aspects are independently evaluated. And we show a screenshot of the annotation on AMT in Figure~\ref{AMT}}.

\noindent\textbf{Results} The manual evaluation results are shown in Table \ref{manual-eva}. To measure the inter-annotator agreement, we calculated Fleiss' kappa~\cite{fleiss1971measuring} for each pair-wise comparison and all the results show fair agreement ($0.2\leq\kappa\leq0.4$) or moderate agreement ($0.4\leq\kappa\leq0.6$). We also conducted sign test to check the significance of the differences. The results indicate that our model performs significantly better than other baselines in both metrics. More specifically, post-training on knowledge bases leads to significant improvements in grammar and logic by offering more knowledge for expanding the story plots. And multi-task learning further enhances the performance in logic and does not affect fluency of generated stories.

\subsection{Relation Understanding}\label{rel_und}
It is still necessary to further investigate whether our model really understands the relations between head and tail entities/events. For example, when our model learns \texttt{car accident \textit{causes} injury} from ConceptNet, it will agree with \texttt{car accident \textit{leads to} injury} and denies \texttt{car accident \textit{is driven by} injury} if our model can identify the specific relation between the head~(\texttt{car accident}) and tail~(\texttt{injury}). By contrast, the model will not distinguish the three statements if it only learns simple relevance (or, co-occurrence) between \texttt{car accident} and \texttt{injury} instead of the specific causal relation.

Therefore, 
we constructed two sets of sentences including \textbf{correct} and \textbf{wrong} knowledge respectively based on the test set of ConceptNet. Specifically, the correct sentences are produced with a {synonymous} template whose relation tokens are replaced by synonyms~(e.g., \texttt{causes} can also be translated to \texttt{leads to}), while the wrong sentences with a {random} template whose relation tokens are randomly replaced by another one. 
Besides, we use \textbf{\textit{training}} template referring to the templates that are used during post-training on knowledge bases. Then, we regard the sentence with lower perplexity as more reasonable. We calculate the accuracy of relation ranking as the percentage of cases where the sentence with wrong template has the highest perplexity compared with the sentences with correct and training templates. 
Furthermore, we also conducted an automatic pair-wise comparison to distinguish the reasonable sentences from unreasonable ones based on the perplexity scores of different models. 
\begin{table}[!ht]
\scriptsize
\centering
\begin{tabular}{lccccc}
\toprule
\multirow{2}{*}{\textbf{Models}}& \multirow{2}{*}{\textbf{Acc}}&\multicolumn{3}{c}{\textbf{Comparison Pairs}}\\
&&\textbf{\textbf{C vs. W}}&\textbf{\textit{T} vs. C} & \textbf{\textit{T} vs. W} \\
\midrule
\textbf{GPT-2~(Pretrain)}& 39.28&53.83 & 44.74& 49.87\\
\textbf{GPT-2(Fine-tune)}& 47.48 &60.31 &39.57 &56.01\\
\midrule
\textbf{Ours} & 67.07 & 71.91& 55.76& 79.89\\
\textbf{~~w/o Knowledge} & 48.07 & 62.07& 42.43& 55.64\\

\bottomrule
\end{tabular}
\caption{Accuracy~(Acc, \%) of relation ranking and winning rates~(\%) of pair-wise comparisons which require selecting a more reasonable sentence from two candidates, each from Correct~(\textbf{C}),  Wrong~(\textbf{W}), or \textit{Training}~(\textbf{\textit{T}}) templates.}
\label{analysis_rel}
\end{table}

    \begin{table*}[!ht]
    \footnotesize
    \centering
    \begin{tabular}{p{2.6cm} p{12.5cm}}
    \toprule
    \textbf{Context:} & [MALE] was \textbf{\underline{driving}} around in the \underline{\textbf{snow}}.\\
    \textbf{Golden Story:}& He needed to \textbf{\underline{get home}} from work. He was driving slowly to \textbf{\underline{avoid accidents}}. Unfortunately the roads were too \textbf{\underline{slick}} and [MALE] \textbf{\underline{lost control}}. His tires \textbf{\underline{lost traction}} and he \textbf{\underline{hit a tree}}. \\
    \midrule
    \textbf{ConvS2S:} & He did n't feel like driving on his way out of the street. He put on his gloves and drove away. [MALE] made a lot of \textbf{\textit{paperwork}} and \textbf{\textit{took a taxi to the mechanic}}. He gave [MALE] a few thousand dollars to \textbf{\textit{help him get up}}.\\
    \textbf{Fusion:} & He was driving for a bit but was getting icy. [MALE] ended up driving over a \textbf{\textit{cliff}}. [MALE] \textbf{\textit{was in the middle of the road}}. He was out of the snow by the time he got home. \\
    \textbf{Plan\&write:} & He noticed a man sitting on the side of the road. He went to the nearest gas station. The man was n't sure if he could \textbf{\textit{have the money}}. The man was \textbf{\textit{left a big deal}}.\\
    \textbf{SKRL:} &A drop of battery stacked in the quarter. [MALE] went to check the phone and turned back. He finally got a car hit off from his way to \textbf{\textit{play again}}. [MALE] had forgotten his \textbf{\textit{hair}}.\\
    \textbf{DSRL:} & A man came in a cop. He called for an ambulance. He said he might go to a \textbf{\textit{park}}. He finally showed up to \textbf{\textit{post what was wrong}}. \\
    \midrule
    \textbf{GPT-2~(Scratch):} & He noticed a large snow pile. He \textbf{\textit{got out}} and \textit{\textbf{began driving}}. Then he {{sped off and went home}}. [MALE] was happy that he had \textit{\textbf{saved his money}}.\\
    \textbf{GPT-2~(Pretrain):} & The doctors \textbf{\textit{took him to a hospital}}. They told him he was going to be fine. He got all the doctors and nurses. The [MALE] parents said \textbf{\textit{they have no hope}}.\\
    \textbf{GPT-2~(Fine-tune):} & He noticed a snowglobe blocking his road. [NEUTRAL] he \textit{\textbf{backed up to the car}} and got a snow shovel. [MALE] used to \textbf{\textit{shovel the snow}} and \textit{\textbf{shovel the snow}}. Now the snow is gone \textit{\textbf{but he has snow boots and snow gloves}.}\\
    \midrule
    \textbf{Ours:} & 
    Suddenly his car \textbf{\underline{broke down}} on the side of the road. [MALE] had to \textbf{\underline{call a tow truck}}. The tow truck \textbf{\underline{came and took [MALE] home}}. [MALE] was happy he was able to \textbf{\underline{get home}}.\\
    \textbf{~~w/o~Pretrain:} &He tried to stop to see how he could get out of his car. His car started to sputter down on him. [MALE] \textbf{\textit{pulled over and kept driving}} . He was able to make up and the car sped away. \\ 
    \textbf{~~w/o~Knowledge:} & He noticed a young lady running off. He stopped and picked her up. When he checked his \textit{\textbf{id}} it was lost. [MALE] realized he had forgotten his \textit{\textbf{id}}.\\
    \textbf{~~w/o~Multi-task:} & He noticed a car in the road . He decided to stop . He \textit{\textbf{got out of his car}}. He \textit{\textbf{drove for half an hour}}.\\
    \bottomrule
    \\
    \end{tabular}
    \caption{Generated stories from different models. \textbf{Bold} words denote the \textbf{key} entities/events in the story. And \textit{italic} words denote the \textit{improper} entities/events in terms of logic and coherence in the context while the \underline{underlined} words are the \underline{proper} ones.}
    \label{cases}
    \end{table*}

{As shown in Table~\ref{analysis_rel}, the external knowledge can help our language model distinguish false sentences from true ones with higher accuracy than GPT-2~(Random chance scores 33.3\%). 
Furthermore, our model prefers the correct template compared with the {wrong} one (winning rate of 71.91\%), and has a close preference between the {{training}} and {correct} templates~(winning rate of 55.76\%). By contrast, GPT-2 without any external knowledge relies more on frequency to score relations, and thus can hardly tell the differences among different types of sentences. 
The results indicate that our model can capture not only simple relevance between head and tail entities/events, but also the specific causal relations.}


\begin{figure*}[!ht]
\centering
\includegraphics[width=6.3in]{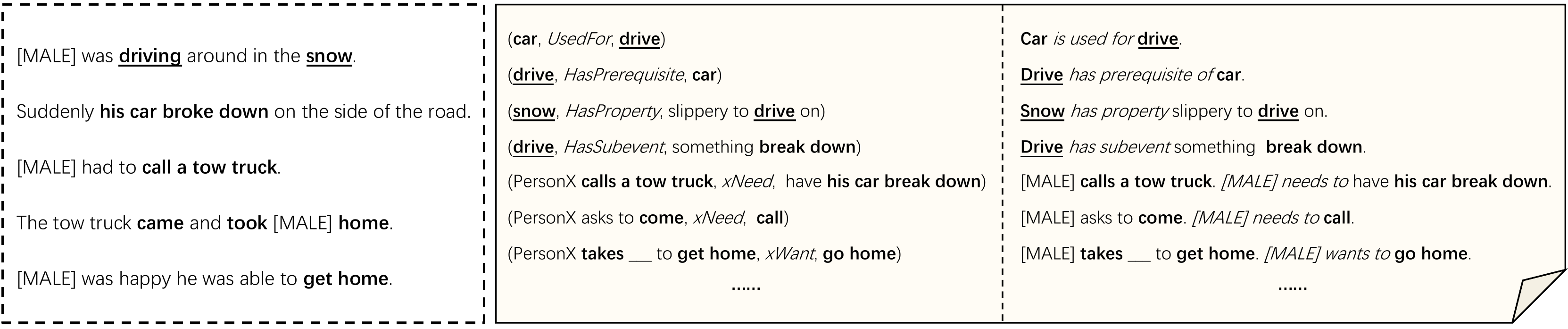}
\caption{An example illustrating how commonsense knowledge facilitates generating reasonable stories. The right block demonstrates interrelated knowledge for the generated story, and the corresponding transformed sentences used in the training. The knowledge is retrieved from ConceptNet and ATOMIC according to the keywords denoted in \textbf{bold}  
in the generated story. And the \underline{underlined} words represent the keywords in the leading context, while the \textit{italic} words represent the relations.}
\label{case_study}
\end{figure*}

\subsection{Case Study}
We presented some generated examples in Table \ref{cases}. Our model can generate more natural and reasonable stories than baselines.

As illustrated, the baselines~(from ConvS2s to DSRL) predict wrong entities and events that are irrelevant to the leading context~(e.g., \texttt{paperwork}), thereby leading to bad overall coherence in the generated stories. Pretrained GPT-2 without any fine-tuning generates an entirely irrelevant, unreasonable story~(e.g., \texttt{hospital}, \texttt{doctor}) due to the lack of knowledge. GPT-2 trained from scratch and fine-tuned GPT-2 suffer from conflicting logic~(e.g., first \texttt{got out} and then \texttt{began driving}, and \texttt{backed up to the car} when \texttt{driving}), repetition~(e.g., \texttt{shovel the snow}), and lousy coherence with some irrelevant keywords~(e.g., \texttt{save money}). In comparison, the story by our model is coherent in logic and fluent in grammar. Furthermore, without pretraining, our model can still incorporate external knowledge to generate a story with an understandable main idea but not always reasonable locally~(e.g., \texttt{pulled over and kept driving}).  When removing knowledge out of our full model, some confusing entities~(e.g., \texttt{id}) will be generated. Besides, removing multi-task learning also significantly affects the logic of generated stories~(e.g., first \texttt{got out} and then \texttt{drove} ) due to the inability of capturing the causal and temporal dependencies in context. 

In order to verify the ability of our model to incorporate external knowledge when generating stories, we showed the utilized commonsense knowledge of this example in Figure \ref{case_study}. We can observe that the external knowledge is useful for expanding a reasonable story plot such as \texttt{driving, broke down, call, came and took home, and get home}.

 \section{Error Analysis}
 Although the proposed model outperforms the state-of-the-art baselines, it needs to be noted that there are still many unreasonable stories losing to other models in manual evaluation. Therefore, we analyzed error types by manually checking all \textit{lost} stories in pair-wise comparisons between our model and two strong baselines including Fusion and GPT-2~(Fine-tune) to reveal the factors that affect the performance. The numbers of stories which lost to our model in logic are 114/102 of 200/200 in total for Fusion/GPT-2~(Fine-tune) respectively. And there are 111 stories of 400 generated by our model losing to these two baselines in logic.
 
 We manually annotated four types of error from the lost stories: {\textbf{repetition}}~(repeating the same scenes), {\textbf{unrelated} entities or events}~(with some wrong keywords but a reasonable main plot), {\textbf{conflicting} logic}~(wrong causal relation or temporal order), and {\textbf{chaotic} scenes}~(difficult to understand). The distribution of different error types is shown in Table \ref{error}. 
 We can observe that unrelated entities/events and conflicting orders make up most of the errors for all the models. Compared with Fusion, GPT-2~(Fine-tune) reduces chaotic scenes effectively but still suffers from severe repetition. Equipped with external knowledge and multi-task learning, our model can further reduce chaotic logic and meanwhile avoid repetition. However, the analysis result illustrates that generating a coherent and reasonable story is challenging.

\begin{table}[!ht]
\scriptsize
\centering
\begin{tabular}{lccc}
\toprule
\multirow{1}{*}{\textbf{Error Type}} & \multirow{1}{*}{\textbf{Ours}} & \multirow{1}{*}{\textbf{Fusion}} & \textbf{GPT-2~(Fine-tune)}\\
\midrule
\textbf{Repetition~(\%)} &1.75 &5.50 & 6.50\\
\textbf{Unrelated~(\%)} &11.25 &16.00 & 15.50\\
\textbf{Conflicting~(\%)} &13.75 &22.00 & 24.50\\
\textbf{Chaotic~(\%)} &1.00 &13.50 & 4.50\\
\bottomrule
\end{tabular}
\caption{Distribution of error types for different models.}
\label{error}
\end{table}
 
\begin{table}[!ht]
\scriptsize
\centering
\begin{tabular}{p{1.4cm} p{5.7cm}}
\toprule
\textbf{Error Type} & \textbf{Cases}\\
\midrule
\textbf{Repetition} &\textbf{[MALE] made up his mind to join the army.} \textit{He was determined to get into the army.} He had never been away from home. \textit{He was determined to get into the army.} He was sent out to Afghanistan.\\
\midrule
\textbf{Unrelated} & \textbf{[MALE] felt he was getting sick.} He had to go to an emergency room. It was his first major surgery. He had a terrible stomach ache. He was nervous about a \textit{test} in an hour.\\
\midrule
\textbf{Conflicting} &\textbf{[FEMALE] swept and mopped the floor.} She put her clothes in the washing machine. She was ready to go to bed. When she was \textit{done}, she \textit{washed the clothes}. She went to bed.\\
\midrule
\textbf{Chaotic} & \textbf{[MALE] was on thin ice with his job.} He had a friend over to help him. [MALE] was able to \textit{hold his breath} the entire time. he was \textit{so cold that he froze }in his tracks. [MALE] finally \textit{felt good} about himself.\\
\bottomrule
\end{tabular}
\caption{Typical errors by our model. \textbf{Bold} sentences are the leading context. \textit{Italic} words denote the improper entities/events in terms of logic and coherence in the context.}
\label{error_cases}
\end{table}

We also presented some typical cases by our model for each error type in Table~\ref{error_cases}. These cases show our model still does not completely prevent logical errors including sentence-level repetition~(\texttt{get into the army}), unrelated entities to the context~(\texttt{test} is obviously unrelated to \texttt{surgery} and \texttt{stomach ache}), conflicting events~(first \texttt{done} but then \texttt{washed the clothes}), and chaotic logic~(due to lack of knowledge about \texttt{on thin ice}). These errors also indicate external knowledge, causal relationships and temporal dependencies play a central role in commonsense story generation.

\section{Conclusions and Future Work}
We present a knowledge-enhanced pretraining model with multi-task learning for commonsense story generation. The proposed framework leverages the implicit knowledge from deep pretrained language models as well as the explicit knowledge by post-training on external commonsense knowledge bases, which leads to better performance for commonsense story generation. 
Besides, in order to further capture the causal and temporal dependencies between the sentences in a story, we employ an auxiliary classification task to distinguish true and auto-constructed fake stories. Extensive experiments show that the proposed method can outperform strong baselines. Further analysis demonstrates that the generated stories are more coherent and reasonable thanks to the use of commonsense knowledge and multi-task learning.

As future work, it would be very interesting to make generative pretraining models have commonsense knowledge without any fine-tuning, namely, integrating the knowledge at the pretraining stage.

\section*{Acknowledgments}

This work was supported by the National
Science Foundation of China (Grant No.
61936010/61876096) and the National Key R\&D
Program of China (Grant No. 2018YFC0830200).
We would like to thank THUNUS NExT Joint-Lab
for the support. We would also like to thank our
action editor, Noah Smith, and the anonymous
reviewers for their invaluable suggestions and
feedback.\\

\bibliography{main-1886-Guan}
\bibliographystyle{acl_natbib}

\end{document}